\documentclass{article}

\usepackage[dblblindworkshop, final]{neurips_2026}

\workshoptitle{The 6th Workshop on Mathematical Reasoning and AI}

\usepackage[utf8]{inputenc}
\usepackage[T1]{fontenc}

\usepackage{hyperref}
\usepackage{url}
\usepackage{booktabs}
\usepackage{amsfonts}
\usepackage{amsmath}
\usepackage{amssymb}
\usepackage{microtype}
\usepackage{graphicx}
\hypersetup{pdftitle={Cost-Effective Automated Judging of Natural-Language Mathematical Proofs},pdfauthor={Benjamin Grayzel}}

\title{Cost-Effective Automated Judging of\\Natural-Language Mathematical Proofs}

\author{%
  Benjamin Grayzel\thanks{Work begun while at the Department of Computer Science, Dartmouth College.}\\
  Independent Researcher\\
  \texttt{benjamingrayzel@gmail.com}
}

\begin{document}

\maketitle

\begin{abstract}
  Grading natural-language mathematical proofs is a recurring cost in
  evaluating math-reasoning systems, and frontier LLM judges are expensive. We
  ask whether cheap open-weight models can serve as reliable judges given a
  candidate proof, a ground-truth proof, and a human-grading rubric. On a
  200-instance validation sample of IMO-GradingBench, two of three cheap judges
  (GPT-OSS-120B, DeepSeek-V4-Flash) and their three-model consensus are
  statistically no worse than the frontier (Claude Opus 4.7, Gemini 3.1 Pro) on agreement with human pass/fail decisions, at
  $4$--$100\times$ lower cost. On the full 1000-instance benchmark, the choice of consensus rule
  over the three judges is a precision/recall dial: unanimous (all-three-pass)
  rules reach the highest precision ($0.855$), majority vote the highest
  recall ($0.912$); across four replicate runs the unanimous rule is also the steadiest. 
  No rule won outright; the dial replicated on a held-out 600-instance split and on the independent ProofBench. 
  In this domain, cheap judges are competitive with the frontier at one to two orders of magnitude lower cost,
  and unanimity is the right setting when false positives are costly.
\end{abstract}

\section{Introduction}
\label{sec:intro}

Benchmarks for AI mathematical reasoning increasingly include problems whose
solutions are full natural-language proofs rather than short final answers, and
scoring those proofs is a bottleneck. Formal verification with proof assistants
such as Lean \citep{moura2021lean} gives trustworthy guarantees and has reached
IMO-medal performance with AlphaProof \citep{hubert2025alphaproof}, but until its
coverage extends to the breadth of research mathematics, most proofs will still
be graded in natural language. While the model under study changes between experiments, the judge
must be reliable and held fixed across an entire study, so its cost is a tax on
the whole research effort, and it recurs inside self-improving systems (which
need a judgment at every step of a generate--critique--revise loop). This
motivates a concrete question: \emph{is there a cheap judge a
budget-constrained researcher can trust for proof grading?}

LLM-as-a-judge \citep{zheng2023judging} is now the standard tool but exhibits
well-documented biases \citep{gu2024survey}, and frontier models often fail to
produce valid arguments even when their final answers are correct
\citep{petrov2025proof}, motivating dedicated grading benchmarks such as the
Open Proof Corpus \citep{dekoninck2025opc} and IMO-GradingBench
\citep{luong2025robust}. Cost has received less attention:
\citet{verga2024poll} show that small-model panels can outperform a single
large judge on QA and chatbot tasks. Two concurrent works frame our
contribution. \citet{ma2026proofgrader} reach expert-level agreement by
combining a strong reasoning backbone with reference solutions and marking
schemes; we ask whether the backbone itself must be strong and find that, for
reference-based pass/fail grading, it need not be. \citet{naik2026frontier}
study the closest reference-\emph{free} version of our question and report
cheap judges trailing the frontier by ${\sim}10\%$ in accuracy; in our
reference-\emph{based} setting the accuracy gap closes entirely, consistent
with the reference solution doing work the judge would otherwise have to do.

We study judging, not solving, on \textbf{IMO-GradingBench}
\citep{luong2025robust}: 1{,}000 instances, each pairing an Olympiad problem
and a reference solution with a candidate proof and an expert human grade on
the standard 0--7 IMO scale. Our prior expectation, following
\citet{verga2024poll}, was that a majority-vote consensus of several cheap
models would be the safest choice. We tested this: the consensus performed
well but did not beat its strongest member. The more useful finding is that
the cheap tier as a whole is competitive: the best cheap open-weight judges are
statistically no worse than frontier baselines (Claude Opus 4.7, Gemini 3.1 Pro) on
pass/fail agreement at a \textbf{cost one to two orders of magnitude lower}. The choice of consensus rule
is most useful as a precision/recall dial whose unanimous setting delivers the highest
precision and the most stable behavior across replicate runs.

\section{Setup}
\label{sec:setup}

\paragraph{Task and metrics.}
A grading instance is a tuple \emph{(problem, ground-truth solution, candidate
solution, human score)}; a judge reads the first three and outputs a score. The
decision that matters most for downstream use is the \textbf{pass/fail
boundary} (a score of $\geq 6$ on the 0--7 IMO scale). Our primary metric is
\textbf{pass-agreement}: the fraction of instances where the judge's pass/fail
decision matches the human's. Because only ${\sim}30\%$ of instances are
passes, a \emph{fail-everything} baseline already reaches $0.715$
pass-agreement on our validation sample, so we report Cohen's $\kappa$
\citep{cohen1960kappa} alongside it (the same agreement corrected for chance,
$0$ for any constant classifier) and include the baseline as a row in every
table. We further report F1 at the pass boundary, Spearman rank correlation with the
human score as a secondary ordinal measure, and per-grading cost.

\paragraph{Data and sampling.}
From the benchmark's 1{,}000 instances (spanning 30 IMO-style problems) we drew
two disjoint random samples of 200: a \emph{prior} (exploratory) sample used
for exploration and for selecting the consensus trio, and a \emph{validation}
sample used as a clean held-out test. The trio, its settings, and the majority
rule were fixed on the prior sample alone, before any validation call was made. Section~\ref{sec:consensus} adds the
cheap tier over the complete benchmark and a check on the 600 instances
untouched during any exploration; headline results are on the validation
sample unless noted. The 200 validation instances come from only 30 problems, so we also
check every paired comparison by resampling whole problems
(Appendix~\ref{app:robust}).

\paragraph{Judges.}
We evaluate three cheap open-weight models (\textbf{GPT-OSS-120B},
\textbf{DeepSeek-V4-Flash}, \textbf{Gemma-4-31B}) and two frontier baselines
(\textbf{Claude Opus 4.7}, \textbf{Gemini 3.1 Pro}). The cheap models were
chosen on the prior sample for their cost-accuracy tradeoff and offsetting
calibration biases (Gemma over-credits, DeepSeek under-credits), the intended
ingredient for a majority vote. Every judge uses the same prompt, lightly
adapted from the benchmark's public grading rubric, and emits a score in
$\{0, 1, 6, 7\}$; each model runs at the strongest reasoning configuration it
exposes (settings, providers, coverage, re-run policy, and the ordinal statistic
behind consensus-row $\rho$ in Appendix~\ref{app:method}). The majority
pass/fail vote of the three cheap models was the pre-specified consensus rule;
Section~\ref{sec:consensus}'s unanimous variants were added later and checked
on held-out data.

\section{Cheap judges are competitive with the frontier}
\label{sec:frontier}

\begin{figure}[t]
  \centering
  \includegraphics[width=0.48\textwidth]{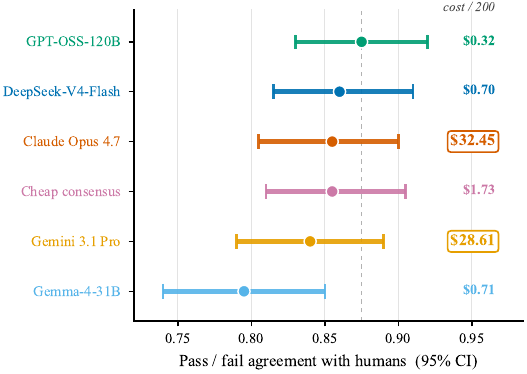}
  \caption{Pass/fail agreement with humans (95\% CI, $n=200$) with per-200-call
    cost at right (frontier costs boxed). Every cheap judge but Gemma-4-31B has
    an interval overlapping the leader's estimate, at one to two orders of
    magnitude lower cost than the frontier baselines.}
  \label{fig:forest}
\end{figure}

Table~\ref{tab:main} and Figure~\ref{fig:forest} report the five judge head-to-head validation
results sorted by pass-agreement, with 95\% bootstrap confidence intervals.

\begin{table}[t]
  \caption{Validation results ($n=200$), sorted by pass-agreement of the
    original run. \emph{4-run} is the mean pass-agreement over four
    independent runs (Appendix~\ref{app:stability}); frontier baselines were
    run once. $\kappa$ is Cohen's kappa at the pass boundary; precision,
    recall, and confusion matrices are in the supplementary material.}
  \label{tab:main}
  \begin{center}
    \begin{small}
      \setlength{\tabcolsep}{3.5pt}
      \begin{tabular}{llcccccr}
        \toprule
        Judge & Reasoning & Pass-agree [95\% CI] & 4-run & $\kappa$ & F1 & Spearman $\rho$ & Cost / 200 \\
        \midrule
        GPT-OSS-120B           & \texttt{xhigh}  & 0.875 [0.830, 0.920] & 0.851 & \textbf{0.716} & 0.806 & 0.623 & \textbf{\$0.32} \\
        DeepSeek-V4-Flash      & default         & 0.860 [0.815, 0.905] & 0.881 & 0.664 & 0.763 & 0.660 & \$0.70 \\
        Cheap consensus (trio) & ---             & 0.855 [0.805, 0.905] & 0.874 & 0.673 & 0.779 & 0.654 & \$1.73 \\
        Claude Opus 4.7        & adaptive        & 0.855 [0.805, 0.900] & ---   & 0.680 & 0.785 & \textbf{0.715} & \$32.45 \\
        Gemini 3.1 Pro         & \texttt{high}   & 0.840 [0.790, 0.890] & ---   & 0.645 & 0.761 & 0.704 & \$28.61 \\
        Gemini 3.1 Pro         & default         & 0.840 [0.790, 0.890] & ---   & 0.655 & 0.771 & 0.714 & \$7.07 \\
        Gemma-4-31B            & \texttt{high}   & 0.795 [0.740, 0.850] & 0.810 & 0.568 & 0.717 & 0.676 & \$0.71 \\
        \addlinespace
        Fail-everything        & ---             & 0.715 [0.650, 0.775] & ---   & 0.000 & 0.000 & ---   & \$0.00 \\
        \bottomrule
      \end{tabular}
    \end{small}
  \end{center}
  \vskip -0.12in
\end{table}

\paragraph{The cheap tier is competitive with the frontier.}
The intervals of the top five systems overlap substantially; paired
comparisons against each frontier baseline make the claim precise. Relative to
Claude Opus 4.7, the paired difference in pass-agreement is $+0.020$ (90\% CI
$[-0.020, +0.060]$) for GPT-OSS-120B, $+0.005$ $[-0.035, +0.045]$ for
DeepSeek-V4-Flash, and $0.000$ $[-0.030, +0.035]$ for the trio; against Gemini
3.1 Pro the same three are $+0.035$ $[-0.010, +0.080]$, $+0.020$
$[-0.020, +0.060]$, and $+0.015$ $[-0.020, +0.055]$. All three are therefore
\textbf{non-inferior to both frontier baselines} at a $\pm 5$-point margin
\citep{schuirmann1987tost}, and DeepSeek and the trio are fully equivalent to
Opus under TOST; GPT-OSS fails equivalence only on the \emph{upper} side.
Gemma is the exception, measurably behind both ($-0.060$ $[-0.095, -0.030]$ vs
Opus, $-0.045$ $[-0.080, -0.010]$ vs Gemini). This shows the best cheap judges are not
measurably worse than the frontier; it does not show they are better, and our
sample cannot rank the leading systems against one another. The chance-corrected column agrees, with
GPT-OSS at $\kappa = 0.716$ and Opus at $0.680$, both ``substantial''
agreement \citep{landis1977measurement} against $\kappa = 0$ for the
fail-everything baseline.

\paragraph{Robustness of the comparison.}
The frontier baselines were run once, and Table~\ref{tab:main} reports the
cheap tier's original run, which for GPT-OSS was the best of its four runs.
The ``no worse than the frontier'' results hold when we resample whole problems instead of
instances, and when we use each cheap system's average over four runs (e.g.\
GPT-OSS vs Opus becomes $-0.005$, 90\% CI $[-0.043, +0.034]$; the trio
$+0.019$ $[-0.015, +0.057]$). They also hold against Gemini at its default
setting, which scores the same as high reasoning at a quarter of the cost
(Appendix~\ref{app:robust}).

\paragraph{The consensus did not beat its best member.}
The majority vote matched Opus on pass-agreement but fell below its strongest
single member, GPT-OSS-120B, at roughly five times the cost: combining a strong
model with two weaker ones diluted rather than improved the result
(Section~\ref{sec:consensus} asks whether a different rule over the same trio
fares better).

\paragraph{The price gap is large.}
GPT-OSS-120B grades at \$0.0016 per instance, about \textbf{$100\times$ cheaper
than Claude Opus 4.7} (\$0.162) and ${\sim}22\times$ cheaper than Gemini 3.1
Pro at its default setting (\$0.035; ${\sim}90\times$ at high reasoning), while
doing no worse on the pass/fail decision. DeepSeek is
${\sim}10\times$ cheaper than default Gemini, and the three-judge consensus
${\sim}4\times$. Running three judges, or one judge three times, for stability costs
\$0.96--\$2.13 per 200: still $3$--$7\times$ cheaper than default Gemini and
$15$--$34\times$ cheaper than Opus. All costs are what OpenRouter charged us in
May 2026; the open models are served there by third-party hosts, so their
prices may change, for example if you pin a provider. Where the frontier keeps an edge is
rank correlation with the human score (Opus $0.715$, Gemini $0.704$, cheap
judges $0.62$--$0.68$): for a graded quality signal rather than a pass/fail
gate, a frontier judge is still safer (Appendix~\ref{app:correlation}). Reasoning effort, meanwhile, is a
model-specific lever, large for GPT-OSS-120B and negligible for Gemini
(Appendix~\ref{app:reasoning}).

\section{Consensus rules are a precision/recall dial}
\label{sec:consensus}

We graded the three cheap judges over the complete 1000-instance benchmark for
a more stable estimate and to explore consensus configurations beyond the
pre-specified majority vote (Table~\ref{tab:full}). Three observations. First,
\textbf{the single-model leader changes with scale}: GPT-OSS-120B led on the
validation sample, but on the full benchmark DeepSeek-V4-Flash is the strongest
single judge ($0.873$ vs $0.842$)---direct evidence that 200 instances cannot
separate the leaders (Appendix~\ref{app:extras}). Second, \textbf{the choice of
consensus rule trades precision against recall}
(Figure~\ref{fig:prdial}, Appendix~\ref{app:extras}): the unanimous all-three-pass rule reaches the
highest pass-agreement ($0.879$) and precision ($0.855$), majority vote is the
most recall-heavy ($0.912$) and, via its median member score, has the highest
rank correlation of any system ($\rho = 0.765$). That unanimity is stricter is true by definition (every unanimous pass is
also a majority pass); what the data add is the size of the trade ($+0.111$
precision for $-0.135$ recall) and that unanimity loses no pass-agreement for it. The same holds for both-pass
pairs (Table~\ref{tab:pairs}): a pair's strictness is driven by its most
conservative member, so any pairing that includes the under-crediting DeepSeek
inherits high precision. Third, \textbf{the top configurations are too close to call}: all-three-pass
is measurably ahead of GPT-OSS and Gemma, but its lead over DeepSeek ($+0.006$,
95\% CI $[-0.008, +0.021]$) and majority vote ($+0.016$ $[-0.011, +0.044]$) is
within noise.

\begin{table}[t]
  \caption{Full benchmark ($n=1000$), 95\% bootstrap CIs; both-pass pairs are
    in Table~\ref{tab:pairs}.}
  \label{tab:full}
  \begin{center}
    \begin{small}
      \begin{tabular}{lcccccc}
        \toprule
        System & Pass-agree [95\% CI] & $\kappa$ & Prec. & Rec. & F1 & Spearman $\rho$ \\
        \midrule
        DeepSeek-V4-Flash      & 0.873 [0.852, 0.893] & 0.717 & 0.815 & 0.812 & 0.814 & 0.732 \\
        GPT-OSS-120B (\texttt{xhigh}) & 0.842 [0.820, 0.864] & 0.668 & 0.716 & 0.889 & 0.793 & 0.704 \\
        Gemma-4-31B (\texttt{high})   & 0.801 [0.776, 0.826] & 0.604 & 0.640 & 0.953 & 0.766 & 0.741 \\
        Majority vote (trio)   & 0.863 [0.842, 0.883] & 0.711 & 0.744 & 0.912 & 0.819 & \textbf{0.765} \\
        All-three-pass         & 0.879 [0.860, 0.898] & \textbf{0.725} & \textbf{0.855} & 0.777 & 0.814 & 0.733 \\
        Fail-everything        & 0.659 [0.629, 0.689] & 0.000 & --- & 0.000 & 0.000 & --- \\
        \bottomrule
      \end{tabular}
    \end{small}
  \end{center}
  \vskip -0.15in
\end{table}

\paragraph{A held-out check of the unanimous rule.}
The unanimous rules were noticed during analysis, so we
checked them on the 600 instances untouched during all exploration, predicting that all-three-pass would win outright. It did not: all-three-pass
scored $0.873$ pass-agreement, a statistical tie with the other consensus
configurations, which land in a $0.868$--$0.878$ band
(Appendix~\ref{app:heldout}). What \emph{did} replicate is the dial itself:
unanimity again delivered the optimal precision ($0.861$ vs the majority
vote's $0.776$) at the cost of recall. This was further replicated on the independent
ProofBench benchmark (Appendix~\ref{app:proofbench}). We therefore recommend
not all-three-pass as uniquely best but \textbf{unanimity as the
high-precision setting} when false positives are costly, and majority
vote when recall or a graded signal matters.

\paragraph{Run-to-run stability.}
Re-running the cheap tier three more times on validation
(Appendix~\ref{app:stability}) shows single cheap judges moving by ${\sim}2$
points (std) between runs (GPT-OSS's headline $0.875$ is the top of a
$0.83$--$0.88$ range). Part of this is ordinary sampling randomness, and part may come from calls
being routed among a shifting pool of third-party providers. The consensus
rules damp this:
all-three-pass has both the highest mean ($0.902$) and the smallest spread
(std $0.009$ vs $0.022$--$0.024$ for the volatile singles). We did not re-run the frontier baselines, so we do not know their spread;
using the cheap tier's four-run averages leaves Section~\ref{sec:frontier}'s
main result unchanged.

\section{Limitations}
\label{sec:limits}

\textbf{Sample size and intervals.} Even on the full benchmark the leading
systems' intervals overlap and we cannot resolve a single best judge; the
frontier baselines were run on the 200-instance validation sample only
(due to budget constraints) and only once, so the full-benchmark comparison is cheap-tier-only and we do not know how much
the frontier varies between runs.
\textbf{Agreement is not verification.} These judges approximate human grades but cannot certify that a proof is
correct. Even the strictest rule passes some proofs the humans failed (roughly
one in seven of its passes).
\textbf{Output scale.} Judges can only answer 0, 1, 6, or 7, so they cannot give the partial
credit (2--5) that humans often do, limiting rank correlation.
\textbf{Budget-constrained search.} We could not afford a broad sweep, most
notably GPT-5.5 Pro; a small blind GPT-5.5 adjudication of the judge--human
disagreements is in Appendix~\ref{app:adjudication}.
\textbf{Contamination.} GPT-OSS-120B predates IMO-GradingBench by three
months, so it cannot have trained on the benchmark's graded instances; grading
against a provided reference is in any case a distinct task from having
encountered the problem. \textbf{Rule
selection.} The unanimous rules were identified during analysis; the held-out
check above and an independent replication on ProofBench
(Appendix~\ref{app:proofbench}) both show the rule \emph{class} replicating
while rankings within the band do not.
\textbf{Scope.} These results speak to judging, not solving, on IMO-style
mathematics with a ground-truth reference and a human grade; we claim no
transfer to reference-free verification.

\section{Conclusion}
\label{sec:conclusion}

For grading AI-generated natural-language proofs on a budget, cheap open-weight
judges are a credible choice: the best of them do as well as frontier
judges on pass/fail agreement at $1$--$10\%$ of the cost. Neither the intuition that a majority
vote would be best nor our logged prediction that the unanimous variant would
win outright was borne out. What survives our checks is a precision/recall
dial: unanimity buys precision and stability, while majority buys recall and the
best ordinal signal. We stop short of crowning a single model, since the
leader shifts between samples and benchmarks.

{\small
  \bibliography{references}

\begin{thebibliography}{13}
\providecommand{\natexlab}[1]{#1}
\providecommand{\url}[1]{\texttt{#1}}
\expandafter\ifx\csname urlstyle\endcsname\relax
  \providecommand{\doi}[1]{doi: #1}\else
  \providecommand{\doi}{doi: \begingroup \urlstyle{rm}\Url}\fi

\bibitem[Cohen(1960)]{cohen1960kappa}
Jacob Cohen.
\newblock A coefficient of agreement for nominal scales.
\newblock \emph{Educational and Psychological Measurement}, 20\penalty0
  (1):\penalty0 37--46, 1960.

\bibitem[de~Moura and Ullrich(2021)]{moura2021lean}
Leonardo de~Moura and Sebastian Ullrich.
\newblock The {Lean} 4 theorem prover and programming language.
\newblock In \emph{Automated Deduction -- CADE 28}, volume 12699 of
  \emph{Lecture Notes in Computer Science}, pages 625--635. Springer, 2021.

\bibitem[Dekoninck et~al.(2025)Dekoninck, Petrov, Minchev, Balunovi{\'c},
  Vechev, Marinov, Drencheva, Konova, Shumanov, Tsvetkov, Drenchev, Todorov,
  Nikolova, Georgiev, Kalinkova, and Ismoldayev]{dekoninck2025opc}
Jasper Dekoninck, Ivo Petrov, Kristian Minchev, Mislav Balunovi{\'c}, Martin
  Vechev, Miroslav Marinov, Maria Drencheva, Lyuba Konova, Milen Shumanov,
  Kaloyan Tsvetkov, Nikolay Drenchev, Lazar Todorov, Kalina Nikolova, Nikolay
  Georgiev, Vanesa Kalinkova, and Margulan Ismoldayev.
\newblock The open proof corpus: A large-scale study of {LLM}-generated
  mathematical proofs.
\newblock \emph{arXiv preprint arXiv:2506.21621}, 2025.

\bibitem[Gu et~al.(2024)Gu, Jiang, Shi, Tan, Zhai, Xu, Li, Shen, Ma, Liu, Wang,
  Zhang, Wang, Gao, Ni, and Guo]{gu2024survey}
Jiawei Gu, Xuhui Jiang, Zhichao Shi, Hexiang Tan, Xuehao Zhai, Chengjin Xu, Wei
  Li, Yinghan Shen, Shengjie Ma, Honghao Liu, Saizhuo Wang, Kun Zhang, Yuanzhuo
  Wang, Wen Gao, Lionel Ni, and Jian Guo.
\newblock A survey on {LLM-as-a-Judge}.
\newblock \emph{arXiv preprint arXiv:2411.15594}, 2024.

\bibitem[Hubert et~al.(2025)Hubert, Mehta, Sartran, Horv{\'a}th,
  {\v{Z}}u{\v{z}}i{\'c}, Wieser, Huang, Schrittwieser, Schroecker,
  et~al.]{hubert2025alphaproof}
Thomas Hubert, Rishi~S. Mehta, Laurent Sartran, Mikl{\'o}s~Z. Horv{\'a}th,
  Goran {\v{Z}}u{\v{z}}i{\'c}, Eric Wieser, Aja Huang, Julian Schrittwieser,
  Yannick Schroecker, et~al.
\newblock Olympiad-level formal mathematical reasoning with reinforcement
  learning.
\newblock \emph{Nature}, 2025.
\newblock \doi{10.1038/s41586-025-09833-y}.
\newblock AlphaProof.

\bibitem[Landis and Koch(1977)]{landis1977measurement}
J.~Richard Landis and Gary~G. Koch.
\newblock The measurement of observer agreement for categorical data.
\newblock \emph{Biometrics}, 33\penalty0 (1):\penalty0 159--174, 1977.

\bibitem[Luong et~al.(2025)Luong, Hwang, Nguyen, Ghiasi, Chervonyi, Seo, Kim,
  Bingham, Lee, Mishra, Zhai, Hu, Michalewski, Kim, Ahn, Bae, Song, Trinh, Le,
  and Jung]{luong2025robust}
Thang Luong, Dawsen Hwang, Hoang~H. Nguyen, Golnaz Ghiasi, Yuri Chervonyi,
  Insuk Seo, Junsu Kim, Garrett Bingham, Jonathan Lee, Swaroop Mishra, Alex
  Zhai, Huiyi Hu, Henryk Michalewski, Jimin Kim, Jeonghyun Ahn, Junhwi Bae,
  Xingyou Song, Trieu~Hoang Trinh, Quoc~V. Le, and Junehyuk Jung.
\newblock Towards robust mathematical reasoning.
\newblock In \emph{Proceedings of the 2025 Conference on Empirical Methods in
  Natural Language Processing (EMNLP)}, pages 35418--35442, Suzhou, China,
  2025. Association for Computational Linguistics.
\newblock Introduces IMO-Bench, including IMO-GradingBench and the four-bucket
  grading prompt.

\bibitem[Ma et~al.(2026)Ma, Cojocaru, Kolhe, Louie, Sharif, Zhang, Zhuang,
  Zaharia, and Min]{ma2026proofgrader}
Wenjie Ma, Andrei Cojocaru, Neel Kolhe, Bradley Louie, Robin~Said Sharif,
  Haihan Zhang, Vincent Zhuang, Matei Zaharia, and Sewon Min.
\newblock Reliable fine-grained evaluation of natural language math proofs.
\newblock In \emph{International Conference on Learning Representations
  (ICLR)}, 2026.
\newblock Introduces ProofBench and ProofGrader.

\bibitem[Naik et~al.(2026)Naik, Shabadi, Alur, and Naik]{naik2026frontier}
Aaditya Naik, Guruprerana Shabadi, Rajeev Alur, and Mayur Naik.
\newblock Do we need frontier models to verify mathematical proofs?
\newblock \emph{arXiv preprint arXiv:2604.02450}, 2026.

\bibitem[Petrov et~al.(2025)Petrov, Dekoninck, Baltadzhiev, Drencheva, Minchev,
  Balunovi{\'c}, Jovanovi{\'c}, and Vechev]{petrov2025proof}
Ivo Petrov, Jasper Dekoninck, Lyuben Baltadzhiev, Maria Drencheva, Kristian
  Minchev, Mislav Balunovi{\'c}, Nikola Jovanovi{\'c}, and Martin Vechev.
\newblock Proof or bluff? evaluating {LLMs} on 2025 {USA} math olympiad.
\newblock \emph{arXiv preprint arXiv:2503.21934}, 2025.

\bibitem[Schuirmann(1987)]{schuirmann1987tost}
Donald~J. Schuirmann.
\newblock A comparison of the two one-sided tests procedure and the power
  approach for assessing the equivalence of average bioavailability.
\newblock \emph{Journal of Pharmacokinetics and Biopharmaceutics}, 15\penalty0
  (6):\penalty0 657--680, 1987.

\bibitem[Verga et~al.(2024)Verga, Hofst{\"a}tter, Althammer, Su, Piktus,
  Arkhangorodsky, Xu, White, and Lewis]{verga2024poll}
Pat Verga, Sebastian Hofst{\"a}tter, Sophia Althammer, Yixuan Su, Aleksandra
  Piktus, Arkady Arkhangorodsky, Minjie Xu, Naomi White, and Patrick Lewis.
\newblock Replacing judges with juries: Evaluating {LLM} generations with a
  panel of diverse models.
\newblock \emph{arXiv preprint arXiv:2404.18796}, 2024.

\bibitem[Zheng et~al.(2023)Zheng, Chiang, Sheng, Zhuang, Wu, Zhuang, Lin, Li,
  Li, Xing, Zhang, Gonzalez, and Stoica]{zheng2023judging}
Lianmin Zheng, Wei-Lin Chiang, Ying Sheng, Siyuan Zhuang, Zhanghao Wu, Yonghao
  Zhuang, Zi~Lin, Zhuohan Li, Dacheng Li, Eric~P. Xing, Hao Zhang, Joseph~E.
  Gonzalez, and Ion Stoica.
\newblock Judging {LLM-as-a-Judge} with {MT-Bench} and {Chatbot Arena}.
\newblock In \emph{Advances in Neural Information Processing Systems 36
  (NeurIPS), Datasets and Benchmarks Track}, 2023.

\end{thebibliography}
  \bibliographystyle{plainnat}
}

\appendix

\section{Methodology details}
\label{app:method}

\paragraph{Judge prompt and scoring buckets.}
Every judge uses the same prompt and is instructed to emit a score in
$\{0, 1, 6, 7\}$ (incorrect / partial / almost / correct); our parser accepts
any integer 0--7, and a handful of off-bucket scores (4 across all runs, under
0.1\%) occurred and are scored as parsed. This four-bucket scheme follows the
public grading prompt released with IMO-GradingBench \citep{luong2025robust},
which we lightly adapt so that our judges are scored under an established,
externally defined rubric rather than one of our own design. Pass/fail and all
confusion-matrix metrics use the raw human score, so a human-4 graded as 6 by a
judge is correctly counted as a false positive. The benchmark also ships a
per-problem marking scheme; we do not provide it to the judge. The exact prompt
is included in the supplementary material.

\paragraph{Sampling.}
The two disjoint 200-instance samples were drawn by uniform selection without
replacement: the prior sample with seed 42, and the validation sample with seed
7 from the remaining 800 instances. The held-out check of
Appendix~\ref{app:heldout} uses the 600 instances in neither sample.

\paragraph{Reasoning settings.}
For each model we use the strongest reasoning configuration it exposes. These
settings are not normalized and are not directly comparable across providers:
GPT-OSS-120B runs at effort \texttt{xhigh}, Gemma-4-31B and Gemini 3.1 Pro with
reasoning activated (listed as \texttt{high}), while Claude Opus 4.7 (adaptive
thinking) and DeepSeek-V4-Flash (default) self-regulate their reasoning, so we
report them at their default.

\paragraph{Providers, coverage, and re-runs.}
The cheap open-weight judges are served through OpenRouter by a shifting pool
of third-party providers; we did not pin a provider, so a single judge's
pass-agreement can move by a few points between runs
(Appendix~\ref{app:stability}). For a reproducible number, pin a provider and
log the served-provider field. The frontier baselines are first-party and
single-provider; we ran them once and did not measure their run-to-run spread. Under $2\%$ of judge calls failed to return a
parseable score on the first attempt (transient rate limits, or reasoning that
exceeded the 32k-token output ceiling without concluding); we re-ran these and
never substituted fabricated scores. Final coverage is 1000/1000 for each cheap
judge and 200/200 for each frontier baseline.

\paragraph{Consensus ordinal statistic.}
For consensus rows we report Spearman $\rho$ of the natural ordinal
statistic (the median member score for majority vote, the minimum for
unanimous rules) whose threshold at ${\geq}6$ reproduces the rule's pass/fail
decision. (The \emph{mean} of member scores would produce intermediate values
that no single judge can emit under the $\{0,1,6,7\}$ bucket scheme, making its
correlation incomparable to single-judge correlations; the min/median
statistics do not have that defect.)

\paragraph{Calibration biases.}
The trio was chosen on the prior sample alone, where the mean signed score
error (judge $-$ human) is $+0.545$ for Gemma-4-31B (over-credits), $-0.790$ for
DeepSeek-V4-Flash (under-credits), and $+0.170$ for GPT-OSS-120B
(near-calibrated). Pooling the prior and validation samples ($n=400$) gives
$+0.51$, $-0.79$, and $+0.02$; on the validation half alone GPT-OSS sits
slightly negative ($-0.13$). The two large biases are stable
across our replicate runs (Gemma $+0.34$ to $+0.46$, DeepSeek $-0.72$ to
$-0.87$), so the trio's offsetting diversity is a property of the models rather
than of any single run; GPT-OSS stays near zero throughout ($-0.08$ to
$-0.03$).

\section{Reasoning effort is a model-specific lever}
\label{app:reasoning}

To check whether reasoning \emph{budget}, rather than model identity, drives
judge quality, we ran Gemini 3.1 Pro at both its default setting and high
reasoning on the validation sample (Table~\ref{tab:gemini}).

\begin{table}[h]
  \caption{Gemini 3.1 Pro: default vs.\ high reasoning ($n=200$). ``Agree'' is
    pass-agreement, $\rho$ is Spearman, and cost is per 200 gradings.}
  \label{tab:gemini}
  \begin{center}
    \begin{small}
      \begin{tabular}{lccccr}
        \toprule
        Setting & Reason.~tok. & Agree & F1 & $\rho$ & Cost \\
        \midrule
        Default & ${\sim}2{,}400$  & 0.840 & 0.771 & 0.714 & \$7.07 \\
        High    & ${\sim}10{,}400$ & 0.840 & 0.761 & 0.704 & \$28.61 \\
        \bottomrule
      \end{tabular}
    \end{small}
  \end{center}
\end{table}

Raising Gemini's reasoning roughly fourfold left its pass-agreement unchanged
($0.840$ in both cases; a paired bootstrap finds no detectable difference, and
ten individual decisions flipped but canceled out), while quadrupling its cost.
By contrast, reasoning effort matters a great deal for GPT-OSS-120B: on the
exploratory sample its pass-agreement rises from $0.72$ at minimal effort to
$0.84$ at \texttt{xhigh}. Claude Opus 4.7 uses adaptive thinking and does not
accept a manual reasoning budget, so its reported result already reflects
whatever thinking it elects to do. The conclusion is that reasoning effort is a
model-specific lever, not a universal one: large for GPT-OSS-120B, negligible
for Gemini.

\section{Run-to-run variance (validation, four runs)}
\label{app:stability}

\begin{figure}[h]
  \centering
  \includegraphics[width=0.6\textwidth]{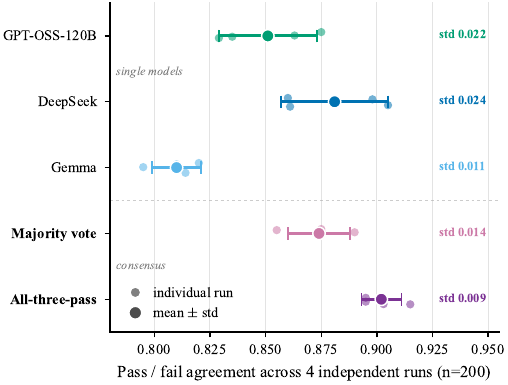}
  \caption{Run-to-run pass-agreement across four independent runs ($n=200$).
    Faded points are individual runs; solid points are mean $\pm$ std.
    All-three-pass is both the highest and the tightest; single cheap judges
    drift more than the consensus rules.}
  \label{fig:runstability}
\end{figure}

We re-ran the cheap tier on the same 200-instance validation sample three more
times, with no fixed random seed, giving four independent runs per system
(including the original). Table~\ref{tab:runstab} reports pass-agreement for
each run; Figure~\ref{fig:runstability} plots the per-run points and their
spread.

\begin{table}[h]
  \caption{Run-to-run pass-agreement on validation ($n=200$).
    \emph{orig}--\emph{rep3} are four independent runs; \emph{mean} and
    \emph{std} are over those four. \emph{Self-maj.} and \emph{self-all-3}
    apply the consensus rules to one model's own three replicates (rep1--3),
    over the problems where all three are valid ($n=191$--$199$): self-maj.\
    passes if $\geq 2$ of 3 runs pass, self-all-3 if all 3 pass.}
  \label{tab:runstab}
  \begin{center}
    \begin{small}
      \setlength{\tabcolsep}{4.5pt}
      \begin{tabular}{lcccccccc}
        \toprule
        System & orig & rep1 & rep2 & rep3 & mean & std & Self-maj. & Self-all-3 \\
        \midrule
        DeepSeek-V4-Flash      & 0.860 & 0.861 & 0.898 & 0.905 & 0.881 & 0.024 & 0.901 & 0.895 \\
        GPT-OSS-120B \texttt{xhigh} & 0.875 & 0.863 & 0.829 & 0.835 & 0.851 & 0.022 & 0.847 & 0.888 \\
        Gemma-4-31B \texttt{high}   & 0.795 & 0.820 & 0.810 & 0.814 & 0.810 & 0.011 & 0.824 & 0.849 \\
        Majority vote (trio)   & 0.855 & 0.875 & 0.890 & 0.875 & 0.874 & 0.014 & --- & --- \\
        \textbf{All-three-pass} & 0.895 & 0.895 & 0.903 & 0.915 & \textbf{0.902} & \textbf{0.009} & --- & --- \\
        \bottomrule
      \end{tabular}
    \end{small}
  \end{center}
\end{table}

Part of this variance is ordinary sampling randomness, which any LLM judge
would show. Another possible contributor is how the models are served: each
call is routed to one of several third-party providers.
Provider dispersion tracks the spread across our three judges: Gemma is served
almost entirely (${\sim}93\%$) by a single provider and has the smallest
spread (std $0.011$, edging out even the majority vote; we do not claim
consensus always reduces variance), while DeepSeek is spread over ten
providers with none serving more than a quarter of calls and has the largest
($0.024$). With only three judges this pattern is suggestive, not conclusive: we log the
provider on every call but did not run a controlled comparison with providers
pinned, so we cannot separate the two causes. For the
recall-biased judges, unanimity over a single model's own three runs filters
the occasional spurious passes that run-to-run noise produces, lifting GPT-OSS from a single-run mean of $0.851$ to $0.888$ and
Gemma from $0.810$ to $0.849$; DeepSeek, already balanced, does not surpass its
best single run. In effect, running one cheap model three times under a
unanimity rule recovers much of the multi-model consensus benefit.

\section{Held-out check of the consensus rules (n=600)}
\label{app:heldout}

The 600 instances in neither the prior nor the validation sample were untouched
during all model selection and rule exploration. Before grading them we logged
the prediction that all-three-pass would outperform the alternative consensus
configurations. Table~\ref{tab:heldout} shows the outcome: it did not. The
consensus configurations land in a $0.868$--$0.878$ band, with all-three-pass
tying DeepSeek+GPT-OSS exactly (identical confusion matrices: because Gemma
over-credits, requiring it to also pass removes nothing that pair had not
already removed). The dial replicates: unanimous rules again give the highest
precision, majority vote the highest recall and F1.

\begin{table}[h]
  \caption{Held-out remaining 600 instances (human pass base rate 221/600).
    The fail-everything baseline scores 0.632 with $\kappa = 0$.}
  \label{tab:heldout}
  \begin{center}
    \begin{small}
      \begin{tabular}{lccccc}
        \toprule
        System & Pass-agree & $\kappa$ & Prec. & Rec. & F1 \\
        \midrule
        DeepSeek-V4-Flash      & 0.878 & 0.737 & 0.843 & 0.824 & 0.833 \\
        GPT-OSS-120B (\texttt{xhigh}) & 0.837 & 0.664 & 0.729 & 0.887 & 0.800 \\
        Gemma-4-31B (\texttt{high})   & 0.807 & 0.622 & 0.662 & 0.973 & 0.788 \\
        DeepSeek + GPT-OSS     & 0.873 & 0.723 & 0.861 & 0.783 & 0.820 \\
        DeepSeek + Gemma       & 0.878 & 0.737 & 0.843 & 0.824 & 0.833 \\
        GPT-OSS + Gemma        & 0.868 & 0.724 & 0.786 & 0.882 & 0.832 \\
        Majority vote (trio)   & 0.873 & 0.738 & 0.776 & 0.923 & 0.843 \\
        All-three-pass         & 0.873 & 0.723 & 0.861 & 0.783 & 0.820 \\
        \bottomrule
      \end{tabular}
    \end{small}
  \end{center}
\end{table}

\section{Independent replication on ProofBench}
\label{app:proofbench}

We replicated the cheap-tier setup on \textbf{ProofBench}
\citep{ma2026proofgrader}: 435 LLM-generated proofs (145 problems from
APMO/EGMO/IMO/Putnam/TST/USAMO; generators o3, Gemini-2.5-Pro,
DeepSeek-R1) with expert grades on the same 0--7 scale, released under MIT
license. Same prompt, same three judges at the same settings, marking schemes
withheld as in our main setup; single run, 435/435 valid scores per judge, no
truncations, \$2.74 total. The expert pass base rate is $94/435 = 0.216$
(vs $0.341$ on IMO-GradingBench), so raw pass-agreement is not comparable
across benchmarks; $\kappa$ is the comparable column.

\begin{table}[h]
  \caption{ProofBench replication ($n=435$; expert pass base rate $0.216$).
    Same rule definitions as Table~\ref{tab:full}.}
  \label{tab:proofbench}
  \begin{center}
    \begin{small}
      \begin{tabular}{lccccc}
        \toprule
        System & Pass-agree & $\kappa$ & Prec. & Rec. & F1 \\
        \midrule
        GPT-OSS-120B (\texttt{xhigh}) & 0.883 & 0.684 & 0.681 & 0.862 & 0.761 \\
        DeepSeek-V4-Flash             & 0.878 & 0.609 & 0.781 & 0.606 & 0.683 \\
        Gemma-4-31B (\texttt{high})   & 0.903 & 0.730 & 0.741 & 0.851 & 0.792 \\
        \addlinespace
        DeepSeek + GPT-OSS     & 0.887 & 0.626 & 0.846 & 0.585 & 0.692 \\
        DeepSeek + Gemma       & 0.897 & 0.656 & 0.877 & 0.606 & 0.717 \\
        GPT-OSS + Gemma        & 0.917 & \textbf{0.754} & 0.815 & 0.798 & 0.806 \\
        \addlinespace
        Majority vote (trio)   & 0.913 & 0.746 & 0.786 & 0.819 & 0.802 \\
        All-three-pass         & 0.894 & 0.644 & \textbf{0.887} & 0.585 & 0.705 \\
        \addlinespace
        Fail-everything        & 0.784 & 0.000 & --- & 0.000 & --- \\
        \bottomrule
      \end{tabular}
    \end{small}
  \end{center}
\end{table}

Three observations. First, the central claim transfers: every cheap judge
again lands in Landis--Koch ``substantial'' territory ($\kappa$
$0.609$--$0.730$, closely tracking $0.604$--$0.717$ on IMO-GradingBench),
against $\kappa = 0$ for a baseline that gets $0.784$ raw agreement from the
skewed base rate alone. Second, \textbf{the dial replicates}: unanimity buys
precision over majority vote ($0.887$ vs $0.786$, $+0.101$) at the cost of
recall ($0.585$ vs $0.819$); on IMO-GradingBench the same trade was
$+0.111$/$-0.135$. Third, \textbf{the single-judge ranking fully reverses}:
Gemma goes from worst to best ($\kappa$ $0.604 \to 0.730$) and DeepSeek from
best to worst ($0.717 \to 0.609$), and all-three-pass---dragged by DeepSeek's
recall collapse---is not the best rule here (the GPT-OSS+Gemma pair and the
majority vote lead on $\kappa$). This is the strongest evidence yet for the
paper's refusal to crown a single judge or rule: what is portable is the trio
plus the dial, not any fixed member or configuration. Caveats: single run
(no replicate spread); judge Spearman $\rho$ is a floor rather than an
estimate here, since $122/435$ expert grades fall in $2$--$5$, values the
$\{0,1,6,7\}$ scheme cannot emit; and we checked for self-preference
(DeepSeek-V4-Flash grading DeepSeek-R1 proofs): it is conservative on all
three generators but least so on its own sibling (judge-minus-expert pass rate
$-0.028$, against $-0.048$ on Gemini-2.5-Pro and $-0.069$ on o3), a small gap
in the direction of self-preference that a sample this size cannot resolve.

\section{Adjudicating judge--human disagreements}
\label{app:adjudication}

The all-three-pass rule disagrees with the human grader on 121 of 1000
instances (45 false positives, 76 false negatives). To probe whether this
residue reflects judge error or instance ambiguity, we had a stronger model
than any judge in our study (GPT-5.5 at \texttt{xhigh} reasoning) blindly
re-grade a seed-fixed sample of 40 disputed instances (25 FP, 15 FN) under the
same prompt, with no access to the human grade or the judges' scores. We
report whose side of the pass/fail boundary it lands on.

\begin{table}[h]
  \caption{Blind GPT-5.5 adjudication of sampled all-three-pass--vs--human
    disagreements.}
  \label{tab:adjudication}
  \begin{center}
    \begin{small}
      \begin{tabular}{lccc}
        \toprule
        Disputed set & $n$ & Sides with judges & Sides with human \\
        \midrule
        False positives (judges pass, human fails) & 25 & 10 (40\%) & 15 (60\%) \\
        False negatives (judges fail, human passes) & 15 & 8 (53\%) & 7 (47\%) \\
        Combined & 40 & 18 (45\%) & 22 (55\%) \\
        \bottomrule
      \end{tabular}
    \end{small}
  \end{center}
\end{table}

Two observations. First, the adjudicator does \emph{not} vindicate the cheap
consensus: it backs the human label slightly more often on both sets. Another model disagreeing with a human does not by itself make the human
label ambiguous; we take the 45\% rate only as a sign that these disagreements
are not mostly clear-cut judge mistakes. The adjudicator's justifications typically reconstruct the argument, accept its broad
strokes, and then turn on one specific technical call (e.g.\ whether citing a
classical theorem without proof is admissible at IMO standard)---exactly the
judgments a human grader makes, which is why a stronger model does not settle
them decisively. Second, much of the false-positive set is a \textbf{scale
artifact}: 32 of the 45 FPs carry a human grade of 4 or 5, values the
$\{0,1,6,7\}$ output scheme cannot emit, so a judge reading a proof as
substantially-right-but-gapped must round to 1 or 6, and 6 crosses the pass
threshold. The adjudicator makes this visible: on the FPs where it sided with
the human it emitted 1 or 0, never 4 or 5 (mean absolute gap to the human
score 2.9 points), so ``siding with the human'' means agreeing on the
threshold while disagreeing sharply on partial credit. Caveats: $n=40$; the
adjudicator ran inside an agentic CLI harness rather than a controlled API
call, so these are indicative observations, not benchmark numbers (one
degenerate generation was re-run; details in the supplementary material).

\section{Robustness of the frontier comparison}
\label{app:robust}

Table~\ref{tab:robust} repeats the paired comparison of
Section~\ref{sec:frontier} in two stricter ways. \emph{Instance} is the
original bootstrap over instances; \emph{cluster} resamples whole problems (the
validation sample has 30), since proofs of the same problem are not
independent; \emph{4-run} uses each cheap system's average over four runs
(Appendix~\ref{app:stability}), again resampling problems. All use $10{,}000$
resamples and a 90\% CI; a system counts as no worse than the frontier
(non-inferior at a 5-point margin) if the lower bound exceeds $-0.05$. Every ``no worse
than the frontier'' result of Section~\ref{sec:frontier} holds under every
variant, including against Gemini at default reasoning, the cheapest frontier
configuration we ran. Two secondary findings are less robust: with wider
intervals, DeepSeek and the trio no longer pass the stricter two-sided
equivalence test against Opus (their upper bounds exceed $+0.05$, i.e.\ they may
be better), and with four-run averages Gemma's gap to Gemini at high reasoning
is no longer distinguishable from zero. On the full benchmark, the same clustered test (95\% CIs) puts
all-three-pass measurably ahead of GPT-OSS ($+0.037$ $[+0.009, +0.065]$) and
Gemma ($+0.078$ $[+0.039, +0.116]$), but not of DeepSeek or majority vote
(Section~\ref{sec:consensus}). Code: \texttt{code/camera\_ready\_checks.py}
in the supplementary material.

\begin{table}[h]
  \caption{Paired difference in validation pass-agreement (system $-$
    frontier), 90\% CIs. \checkmark: no worse than the frontier (lower bound above $-0.05$). Gem.\
    high/def.: Gemini 3.1 Pro at high/default reasoning.}
  \label{tab:robust}
  \begin{center}
    \begin{footnotesize}
      \setlength{\tabcolsep}{2.5pt}
      \begin{tabular}{llccc}
        \toprule
        Reference & System & Instance & Cluster & 4-run, cluster \\
        \midrule
        Opus & GPT-OSS  & $+0.020$ $[-0.020, +0.060]$ \checkmark & $+0.020$ $[-0.024, +0.068]$ \checkmark & $-0.005$ $[-0.043, +0.034]$ \checkmark \\
                 & DeepSeek & $+0.005$ $[-0.035, +0.045]$ \checkmark & $+0.005$ $[-0.037, +0.052]$ \checkmark & $+0.027$ $[-0.013, +0.072]$ \checkmark \\
                 & Trio     & $+0.000$ $[-0.030, +0.035]$ \checkmark & $+0.000$ $[-0.039, +0.043]$ \checkmark & $+0.019$ $[-0.015, +0.057]$ \checkmark \\
                 & Gemma    & $-0.060$ $[-0.095, -0.030]$ & $-0.060$ $[-0.090, -0.030]$ & $-0.046$ $[-0.086, -0.006]$ \\
        \addlinespace
        Gem.\ high & GPT-OSS  & $+0.035$ $[-0.010, +0.080]$ \checkmark & $+0.035$ $[-0.015, +0.089]$ \checkmark & $+0.010$ $[-0.039, +0.058]$ \checkmark \\
                      & DeepSeek & $+0.020$ $[-0.020, +0.060]$ \checkmark & $+0.020$ $[-0.025, +0.068]$ \checkmark & $+0.042$ $[+0.001, +0.086]$ \checkmark \\
                      & Trio     & $+0.015$ $[-0.020, +0.055]$ \checkmark & $+0.015$ $[-0.033, +0.065]$ \checkmark & $+0.034$ $[-0.009, +0.078]$ \checkmark \\
                      & Gemma    & $-0.045$ $[-0.080, -0.010]$ & $-0.045$ $[-0.079, -0.012]$ & $-0.031$ $[-0.070, +0.010]$ \\
        \addlinespace
        Gem.\ def. & GPT-OSS  & $+0.035$ $[-0.005, +0.075]$ \checkmark & $+0.035$ $[-0.005, +0.079]$ \checkmark & $+0.010$ $[-0.031, +0.050]$ \checkmark \\
                         & DeepSeek & $+0.020$ $[-0.025, +0.065]$ \checkmark & $+0.020$ $[-0.023, +0.064]$ \checkmark & $+0.042$ $[+0.003, +0.082]$ \checkmark \\
                         & Trio     & $+0.015$ $[-0.020, +0.050]$ \checkmark & $+0.015$ $[-0.024, +0.056]$ \checkmark & $+0.034$ $[+0.000, +0.070]$ \checkmark \\
                         & Gemma    & $-0.045$ $[-0.075, -0.015]$ & $-0.045$ $[-0.065, -0.026]$ & $-0.031$ $[-0.058, -0.004]$ \\
        \bottomrule
      \end{tabular}
    \end{footnotesize}
  \end{center}
\end{table}

\section{Additional results}
\label{app:extras}

\paragraph{Both-pass pairs.}
Table~\ref{tab:pairs} reports the pairwise unanimous rules on the full
benchmark; adding the over-crediting Gemma to an already-strict pair changes
little (all-three-pass and DeepSeek+GPT-OSS differ on a single instance).

\begin{table}[h]
  \caption{Both-pass pairs, full benchmark ($n=1000$), 95\% bootstrap CIs.
    Consensus $\rho$ uses the minimum member score (Appendix~\ref{app:method}).}
  \label{tab:pairs}
  \begin{center}
    \begin{small}
      \begin{tabular}{lcccccc}
        \toprule
        System & Pass-agree [95\% CI] & $\kappa$ & Prec. & Rec. & F1 & Spearman $\rho$ \\
        \midrule
        DeepSeek + GPT-OSS     & 0.878 [0.859, 0.897] & 0.723 & 0.852 & 0.777 & 0.813 & 0.733 \\
        DeepSeek + Gemma       & 0.877 [0.857, 0.897] & 0.725 & 0.824 & 0.812 & 0.818 & 0.738 \\
        GPT-OSS + Gemma        & 0.866 [0.845, 0.886] & 0.712 & 0.765 & 0.877 & 0.817 & 0.744 \\
        \bottomrule
      \end{tabular}
    \end{small}
  \end{center}
\end{table}

\begin{figure}[h]
  \centering
  \includegraphics[width=0.6\textwidth]{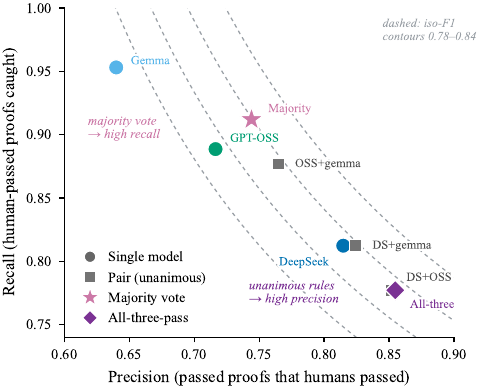}
  \caption{Consensus rule as a precision/recall dial on the full benchmark
    ($n=1000$); dashed lines are iso-F1 contours. Majority vote sits at high
    recall; the unanimous rules move toward high precision.}
  \label{fig:prdial}
\end{figure}

\paragraph{The precision/recall dial.}
Figure~\ref{fig:prdial} plots the consensus configurations of
Table~\ref{tab:full} and Table~\ref{tab:pairs} in precision/recall space.

\begin{figure}[h]
  \centering
  \includegraphics[width=0.6\textwidth]{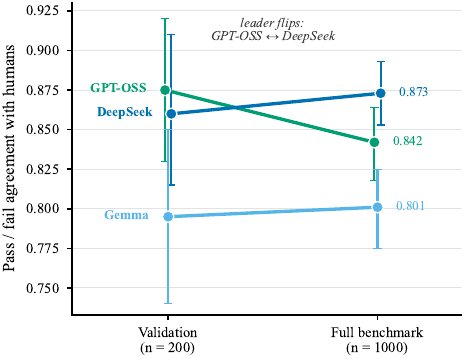}
  \caption{The strongest single judge flips between the validation sample and
    the full benchmark; every interval overlaps.}
  \label{fig:leaderflip}
\end{figure}

\paragraph{Leader flip between samples.}
Figure~\ref{fig:leaderflip} shows the strongest single judge flipping between
the 200-instance validation sample and the full 1000-instance benchmark
(GPT-OSS $\leftrightarrow$ DeepSeek); the 95\% CIs overlap throughout, so 200
instances cannot separate the leaders.

\begin{figure}[h]
  \centering
  \includegraphics[width=0.5\textwidth]{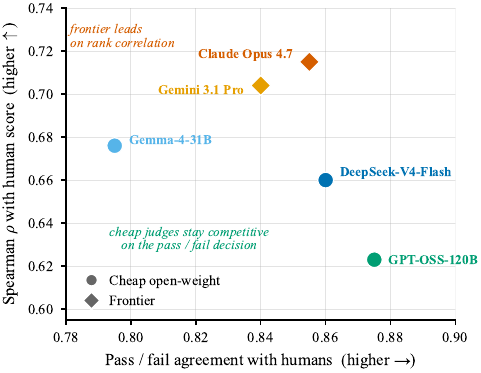}
  \caption{Pass/fail agreement (x) versus Spearman rank correlation with the
    human score (y) on the validation sample ($n=200$). Cheap open-weight
    judges (circles) stay competitive on the pass/fail decision; the frontier
    baselines (diamonds) lead on rank correlation by a modest margin.}
  \label{fig:passrank}
\end{figure}

\paragraph{Pass/fail vs.\ rank correlation.}
\label{app:correlation}
Figure~\ref{fig:passrank} plots pass/fail agreement against Spearman rank
correlation with the human score on the validation sample. The cheap judges
stay competitive on the pass/fail decision, while the two frontier models lead
on rank correlation by a modest margin; DeepSeek-V4-Flash is the exception that
does well on both. This lead is measured on the validation sample only: on the
full benchmark the cheap judges' rank correlation rises to $0.70$--$0.77$
(Table~\ref{tab:full}), matching the frontier's validation values, so the gap
may be a small-sample effect. We did not run the frontier baselines on the full
benchmark, so a matched full-benchmark correlation comparison is not available.

\paragraph{Confusion matrices.}
On the full benchmark the unanimous all-three-pass rule has the fewest false
positives (45/1000) and the recall-biased Gemma the most (183/1000); complete
confusion matrices for every system and sample are in the supplementary
material, along with per-instance scores and the standalone code that
regenerates every number in this paper.

\end{document}